\documentclass[
  pre,
  twocolumn,
  twoside,
  byrevtex,
  superscriptaddress,
  floatfix,
  longbibliography,
]{revtex4-2}

\usepackage[realmainfile]{currfile}
\usepackage{amssymb}
\usepackage{amsmath,booktabs}

\usepackage{url}
\PassOptionsToPackage{hyphens}{url}\usepackage{hyperref}
\makeatletter
\g@addto@macro{\UrlBreaks}{\UrlOrds}
\makeatother

\usepackage{hyperref}
\hypersetup{
  colorlinks=true,
  allcolors=todoblue,
  urlcolor=todoblue,
  citecolor=todoblue,
  pdfborder={0 0 0},
  breaklinks=true,
}

\usepackage{ulem}

\usepackage{colortbl}

\makeatletter

\def\CT@@do@color{%
  \global\let\CT@do@color\relax
  \@tempdima\wd\z@
  \advance\@tempdima\@tempdimb
  \advance\@tempdima\@tempdimc
  \advance\@tempdimb\tabcolsep
  \advance\@tempdimc\tabcolsep
  \advance\@tempdima2\tabcolsep
  \kern-\@tempdimb
  \leaders\vrule
  \hskip\@tempdima\@plus  1fill
  \kern-\@tempdimc
  \hskip-\wd\z@ \@plus -1fill }
\makeatother

\makeatletter
\newcommand*{\centerfloat}{%
  \parindent \z@
  \leftskip \z@ \@plus 1fil \@minus \textwidth
  \rightskip\leftskip
  \parfillskip \z@skip}
\makeatother

\usepackage{xcolor}

\definecolor{olivegreen}{rgb}{0.33333,.41961,0.18431}
\definecolor{forestgreen}{rgb}{0.13333,.5451,0.13333}
\definecolor{lightgrey}{rgb}{0.7,0.7,0.7}
\definecolor{verylightgrey}{rgb}{0.90,0.90,0.90}
\definecolor{veryverylightgrey}{rgb}{0.95,0.95,0.95}
\definecolor{grey}{rgb}{0.5,0.5,0.5}

\definecolor{headerblue}{HTML}{33367E}
\definecolor{unitednationsblue}{HTML}{4D88FF}

\definecolor{charcoal}{HTML}{36454F}
\definecolor{cinerous}{HTML}{98817B}
\definecolor{feldgrau}{HTML}{4D5D53}
\definecolor{glaucous}{HTML}{6082B6}
\definecolor{arsenic}{HTML}{3B444B}
\definecolor{xanadu}{HTML}{738678}

\definecolor{firebrick}{HTML}{B22222}
\definecolor{orangered}{HTML}{FF4500}
\definecolor{tomato}{HTML}{FF6347}

\definecolor{purpletaupe}{HTML}{3B444B}

\usepackage{ulem}

\definecolor{todoblue}{RGB}{0, 91, 187}

\usepackage{graphicx,epsfig,verbatim,enumerate}

\usepackage{ifthen}

\usepackage{longtable}

\usepackage{mathtools}

\newboolean{twocolswitch}

\newcommand{\sindex}[1]{}
\newcommand{\nindex}[1]{}

\newcommand{\etal}{\textit{et al.}}
\newcommand{\www}[1]{\url{#1}}

\newcommand{\anchor}[1]{`\texttt{#1}'}

\usepackage{lettrine}

\setboolean{twocolswitch}{true}

\begin{document}

\title{\protect
  Curating corpora with classifiers:
A case study of clean energy sentiment online
}

\author{
\firstname{Michael V.}
\surname{Arnold}
}
\email{mvarnold@uvm.edu}
\affiliation{
  Computational Story Lab,
  Vermont Complex Systems Center,
  MassMutual Center of Excellence for Complex Systems and Data Science,
  Vermont Advanced Computing Core,
  University of Vermont,
  Burlington, VT, USA
  }

\author{
\firstname{Peter Sheridan}
\surname{Dodds}
}
\affiliation{
  Computational Story Lab,
  Vermont Complex Systems Center,
  MassMutual Center of Excellence for Complex Systems and Data Science,
  Vermont Advanced Computing Core,
  University of Vermont,
  Burlington, VT, USA
  }
\affiliation{
  Department of Computer Science,
  University of Vermont,
  Burlington, VT, USA
}

\author{
\firstname{Christopher M.}
\surname{Danforth}
}
\affiliation{
  Computational Story Lab,
  Vermont Complex Systems Center,
  MassMutual Center of Excellence for Complex Systems and Data Science,
  Vermont Advanced Computing Core,
  University of Vermont,
  Burlington, VT, USA
  }
\affiliation{
  Department of Mathematics \& Statistics,
  University of Vermont,
  Burlington, VT, USA
  }

\date{\today}

\begin{abstract}
  \protect
  Well curated, large-scale corpora of social media posts containing broad public opinion
offer an alternative data source to complement traditional surveys.
While surveys are effective at collecting representative samples and are capable of achieving high accuracy,
they can be both expensive to run and lag public opinion by days or weeks.
Both of these drawbacks could be overcome 
with a real-time, high volume data stream 
and fast analysis pipeline.
A central challenge in orchestrating
such a data pipeline 
is devising an effective method
for rapidly selecting the best corpus of relevant documents for analysis. 
Querying with keywords alone often includes irrelevant documents
that are not easily disambiguated with bag-of-words natural language processing methods. 
Here, we explore methods of corpus curation to filter irrelevant tweets using pre-trained transformer-based models,
fine-tuned for our binary classification task on hand-labeled tweets.
We are able to achieve F1 scores of up to 0.95.
The low cost and high performance of fine-tuning such a model suggests
that our approach could be of broad benefit as a pre-processing step for social media datasets with uncertain corpus boundaries.
 
\end{abstract}


\maketitle

\section{Introduction}
\label{sec:corpusCreation.introduction}


The wide-spread availability of social media data
has resulted in an explosion of social science studies
as researchers adjust
from data scarcity
to abundance
in the digital age~\cite{lazer2009computational,lazer2021meaningful}.
The potential for large scale digitized text to
help understand human behavior remains immense.
Researchers have attempted to quantify myriad social phenomena through changes in language use of societies over time, 
typically through the now massive collections of digitized books and texts~\cite{michel2011quantitative}
or natively digital large-scale social media datasets~\cite{alshaabi2021storywrangler}.

Analysis of social media data promises to supplement
traditional polling methods
by allowing for
rapid, near real-time measurements
of public opinion,
and for historical studies of public language \cite{oconnor2010, cody2015climate, cody2016, wu2023}.
Polling remains the gold standard
for measuring public opinion where precision matters,
such as predicting the outcomes of elections. 
Where trends in attention or sentiment suffice,
social media data can provide insights at dramatically lower costs~\cite{o2010tweets}. However, for targeted studies using social media data, researchers need a principled way to define the potentially arbitrary boundaries of their corpus~\cite{shugars2021pandemics}.

When researchers characterize online discourse around a specific topic,
a few approaches are available.
Each comes with trade-offs,
both in the costs of researchers' time,
as well as the resulting precision and recall of the corpus. 

For some studies a corpus is best defined by a set of relevant users, such as a set of politicians' social media accounts or the set of users following a notable account~\cite{jungherr2016twitter}. 
Studies that observe the behavior of networked publics often take this user-focused approach~\cite{bode2016politics}.
For studies of social media advertising, a list of relevant buyers can be used to define the boundaries, whether politicians or companies~\cite{aisenpreis2023us,lee2018advertising}.

To curate a topic-focused corpus limited keyword filters can be an effective strategy.
Keywords can be used to match a broad cross-section of relevant posts with high precision,
but often have low recall~\cite{llewellyn2015extracting}.
Relevant hashtags can signal a user's intent to join a specific online conversation
beyond their immediate social network.
Hashtag based queries have been used by researchers
to construct focused corpora of tweets ranging from sports and music~\cite{blaszka2012worldseries,choi2014south},
to public health, natural disasters, political activism, and
protests~\cite{lienemann2017methods,steinert2015online,lotan2011arab,freelon2016beyond,jackson2020hashtagactivism,gallagher2019reclaiming,gallagher2018divergent,gorodnichenko2021social,arnold2021hurricanes}.

Alternatively, researchers can query for posts with an expansive set of keywords
to increase recall at the expense of precision. 
Researchers can generate such a set of keywords algorithmically,
or by asking experts with domain knowledge, or via a combination of the two.
Expert-crafted keyword lists have been used by researchers to study topics such as
social movements and responses to the COVID-19 pandemic~\cite{jackson2020hashtagactivism,shugars2021pandemics,chen2020tracking,green2020elusive}.
Other researchers have generated lists of keywords algorithmically, e.g., using Term Frequency - Inverse Document Frequency (TF-IDF)~\cite{aizawa2003information} and word embeddings~\cite{marujo2015automatic}, or
by comparing the distribution of words in a corpus of interest to a reference corpus
and selecting words with high rank-divergence contributions~\cite{dodds2020allotaxonometry,alshaabi2021world,minot2022distinguishing,alajajian2015,stupinski2022quantifying}. 
Regardless of the methods used to choose keywords, 
continued expansion beyond the most relevant necessarily reduces precision. 
Researchers can further refine the set of relevant keywords to balance precision and recall,
and add complexity to their queries with exclusion terms or Boolean operators to require multiple keywords.
The possibilities are endless~\cite{schwartz2000self} 
and reviewers have little information available to decide if the choices made were appropriate.

While some topic-focused social media datasets can be well curated
with simple heuristics
or rules-based classifiers,
others could benefit from an alternative paradigm.
Here, we argue for a two step pre-processing pipeline
that combines broad,
high recall keyword queries
with fine-tuned,
transformer-based classifiers
to increase precision.
Our approach can trade the labor costs associated with building rules-based filters, 
for the cost of labeling social media data, which could potentially be further reduced using few-shot learning~\cite{wang2020generalizing}, 
while still achieving high precision. 

The tools available for text classification have improved significantly over the past decade.
Since the introduction of Word2Vec in 2013 and GloVe in 2014,
the natural language processing community has had access to high quality, global word embeddings~\cite{mikolov2013efficient,pennington2014glove}.
These embeddings are trained vector representations of words from 
a given corpus of text,
enabling word comparisons with distance metrics.
However, global embeddings average the representations of words,
making them unsuitable for document classification
where key terms have multiple meanings.
The subsequent development of large pre-trained language models
enabled high performance on downstream tasks with relatively little additional computational cost to fine-tune~\cite{devlin2018bert,liu2019roberta}.
Such models provide contextual, rather than global, word embeddings. 

Since 2019, pre-trained language models have become less resource intensive while improving performance.
Knowledge distillation has enabled models like DistilBert and MiniLM, 
which retain the performance of full sized models while requiring significantly less memory
and performing inference more rapidly~\cite{sanh2019distilbert,wang2020minilm}. 
Smaller, faster models enable researchers with limited resources to adopt these tools for NLP tasks, requiring only a laptop for state-of-the-art performance.
Improved pre-training, introduced with MPNet,
combines the benefits of masked language modeling (MLM)
and permuted language modeling (PLM),
better making use of available token and position information~\cite{song2020mpnet}.

While transformer-based language models provide state of the art performance on natural language processing tasks,
they can be difficult to understand and visualize. 
Using twin and triplet network structures,
pre-trained models can be trained to generate 
semantically meaningful sentence embeddings that can be compared using cosign distances~\cite{reimers-2019-sentence-bert}. 
Through pre-training with contrastive learning on high quality datasets, general purpose sentence embeddings like E5 have become the new state-of-the-art~\cite{wang2022text}.

Text classification still remains a difficult task. 
existing models are less successful with longer texts~\cite{gao2021limitations},
and text classification with a large number of classes remains challenging~\cite{chang2020taming}. 
However, for the specific task of classifying tweets~\cite{antypas2022twitter}
as `relevant' (R) or `non-relevant' (NR) to a specific topic---an instance of binary classification---we feel existing models are sufficiently capable.
Sophisticated, pre-trained language models are readily accessible to researchers from Hugging Face~\cite{wolf2020transformers}
and can be easily fine-tuned with a limited amount of labeled data~\cite{yan2018few,wang2020generalizing}.
Tools like ChatGPT have been shown to outperform untrained human crowd-workers
for zero-shot text classification, 
while costing an order of magnitude less~\cite{gilardi2023chatgpt}.


As a case study, we examine online language around emission-free energy technologies. 
In democratic societies the social perception of technologies
affects the willingness of governments
to extend subsidies,
expedite permitting,
or regulate competing energy sources, 
ultimately effecting the energy mix of the grid.
Quantifying public attitudes is useful for policy makers to be responsive to public preferences
and for science communicators to respond when public opinion does not reflect expert consensus.

To quantify public perceptions of energy on social media sites,
researchers have use a variety of methods to curate tweet corpora. 
This could be as simple as querying for a single hashtag.
Jain \etal\ choose `\#RenewableEnergy'  to generate a corpus for a renewable energy classification study~\cite{jain2019sentiment}.
Zhang \etal\ query for tweets containing a list of hashtags,
before quantifying overall attention trends and sentiment by energy source~\cite{zhang2022perceptions}.
Li \etal\ use a two-phase approach, querying for relevant hashtags, before filtering non-relevant tweets with keywords, such as those containing both `\#solar' and `eclipse', with filter keywords built on a trial-and-error approach~\cite{li2019beyondBigData}.
Alternatively, Kim \etal\ use keyword phrases,
such as `solar energy' and `solar panel', to search for relevant tweets,
before using RoBERTa to classify sentiment~\cite{kim2021public}.
V{\aa}ger{\"o} \etal\ use a contextual language model to classify sentiment of tweets towards wind power in Norway~\cite{vaagero2023machine}.
Using Reddit, Kim \etal\ study renewable energy discourse by collecting all messages from a particular subreddit,
a page devoted to a topic,
before analyzing a word co-occurrence network~\cite{kim2020exploring}.

Published studies use a wide range of corpus curation techniques
and provide varying levels of justification for each choice.
Although we focus on the topic of renewable energy, we hope our methods are broadly applicable
to any text-based social media dataset.

We structure the remainder of this paper as follows. 
In the Methods and Data section we present a description of our dataset 
and discuss the task of relevance classification as it relates to corpus curation. 
In the Results section, we present case studies for the keywords \anchor{solar}, \anchor{wind}, and \anchor{nuclear}. 
We examine the ambient sentiment time series for each corpus, and compare measurements between the unfiltered, relevant and non-relevant text.
To show the differences in language between these corpora, we present sentiment shift plots~\cite{gallagher2021generalized} and allotaxonographs~\cite{dodds2020allotaxonometry}.
Finally, we share concluding remarks and potential future research.

\section{Methods and Data}
\label{sec:corpusCreation.methods}

We explore the performance of text classifiers
powered by contextual sentence embeddings
for social media corpus curation
through a selection of case studies related to clean energy. 

\subsection{Description of data sets}
\label{sec:corpusCreation.data}

In this study, we examine ambient tweet datasets,
collections of tweets that are anchored by a single keyword or set of keywords. 
From Twitter's Decahose API,
a random 10\% sample of all public tweets,
we select tweets containing user-provided locations~\cite{twitterDecahose}. 
We extracted these locations from a free text location field in each user's bio,
if the text matched a valid 
\texttt{`city, state'} string in the United States~\cite{gray2018english, linnell2021sleep}.
From this selection,
we query for tweets that both contain keywords of choice and are classified as being written in the language English by FastText~\cite{joulin2017bag}.
We define the results of this query as the unfiltered ambient corpus.

To illustrate the utility of our methods,
we chose three keywords related to non-fossil fuel energy generating technologies, \texttt{`wind'}, \texttt{`solar'}, and \texttt{`nuclear'}. 
Over the study period from 2016 to 2022,
these keywords matched 3.43M, 1.39M, and 1.29M tweets in our subsample, respectively. 
In Tab.~\ref{tab:example_tweets}, 
we show example tweets from each corpus.
We binned tweets into windows of two weeks,
balancing the desire for large sample sizes for each bin with the need for higher resolution to show short term dynamics.
While the terms of our service agreement with Twitter do not allow us to publish raw tweets,
we provide relevant tweet IDs for rehydration. 

\subsection{Sentence embeddings}
To better visualize the results of our classification algorithms, we chose pre-trained language models which had been fine-tuned to perform sentence embeddings.
We also considered that vector representations for sentences would better align with our desired abstraction level for the relevance classification task.

\subsection{Relevance classification}
Our task of interest is classifying if a post, 
in its entirety,
is relevant to the researcher's chosen topic of interest. 
Conceptually, this task is related to semantic textual similarity, 
for which sentence embeddings
have achieved state of the art performance~\cite{han2013umbc_ebiquity, chandrasekaran2021evolution}. 
Rather than finding nearest neighbors in a semantic space,
we are training a classifier to partition the semantic space into relevant and non-relevant regions.

For training, we hand-label a random sample of 1000 matching tweets
for each keyword
as either `Relevant' (R)  or `Non-Relevant' (NR) to energy production. 
We have made tweet IDs and corresponding labels available for both the training data 
as well as predicted labels for the full data set.

We then fine-tune nine models for comparison, based on pre-trained contextual sentence embeddings~\cite{song2020mpnet, wang2020minilm}.
We list the performance of these models in Table~\ref{tab:F1-scores}.
For each model we labeled a random sample of one thousand (1,000) tweets.
We choose a train-test split of 67\% and 33\%. 
Tweets are limited to a max of 280 characters for the duration of our study period, 
shorter than the minimum truncation length of 256 word pieces for the models we tested.

\begin{table}[t]
    \begin{tabular}{  l p{1.1cm} p{4.55cm} }
        \toprule
        \textbf{Keyword} & \textbf{Class}     
        & \textbf{Example Tweet}   \\\midrule
    Solar & (R)
            &  The decreasing costs of solar and batteries mean a sustainable future is closer than we think.  \\ \cmidrule(lr){2-3} 
        & (NR)      
            & Looks like there's a solar eclipse down here. The space nerds bought all the hotel rooms.    \\\midrule
    Wind & (R)      
            & At this time of year wind makes up only a fraction of the state's energy generation mix.  \\ \cmidrule(lr){2-3} 
        & (NR)      
            & His mom caught wind of what they were up to and shut down their plans pretty quickly. \\\midrule
    Nuclear & (R)
            & Nuclear activists are questioning \#MAYankee's accelerated decommissioning plan. \\ \cmidrule(lr){2-3} 
        &  (NR)      
            & The global nuclear arsenal stands around 10,000 warheads, down from 70,000 at the peak of the Cold War. \\
        \bottomrule
    \end{tabular}
    \caption{
    \textbf{Paraphrased example tweets for relevant (R) and non-relevant (NR) examples in each case study.}
    To label the training data, we defined relevant tweets as those which are related to the topic of electricity generation or clean energy. Non-relevant tweets contained the keyword, but were wholly or primarily unrelated.
    }
    \label{tab:example_tweets}
\end{table}

\begin{table}[t]
\begin{tabular}{lllll}
\toprule
   & \anchor{solar} & \anchor{wind} & \anchor{nuclear} \\
     \cmidrule(lr){2-4} 
   \% Relevant  & 43.7\% & 4.7\% & 16.0\% \\
   \midrule
    F1 - MPNet                  & 0.951         & \textbf{0.903}  & 0.860  \\
    F1 - MiniLM-L12             & 0.933         & 0.839   & 0.879  \\
    F1 - MiniLM-L6              & 0.949         &  0.828  & 0.857   \\
    F1 - DistilRoberta          & \textbf{0.956}  &  \textbf{0.903}  & 0.857  \\
    F1 - paraphrase-MiniLM-L6   & 0.943         &  0.800  & 0.826  \\
    F1 - paraphrase-MiniLM-L3   & 0.918         &  0.714  & 0.814  \\
    F1 - distiluse-multilingual & 0.929         &  0.759   & \textbf{0.912}  \\
    F1 - e5-base                & 0.949         &  0.867   & 0.881 \\
    F1 - e5-large               & 0.949         &  0.828    & 0.895 \\

  \bottomrule
\end{tabular}
\caption{
  \textbf{Summary statistics and model performance for each of the three case studies.}
  First, we report the proportion of human labeled tweets
that are labeled relevant to clean energy from our thousand tweet subsample.
  The \anchor{solar} corpus is most evenly split,
while the \anchor{wind} corpus is the most imbalanced. 
  Second, we detail F1 evaluation scores for 
  a range of fine-tuned
  text classifiers trained on our labeled data.
  The model performance does not necessarily degrade dramatically
  for corpora with a small proportion of relevant documents, such as for \anchor{wind}.}
\label{tab:F1-scores}
\end{table}

\begin{figure*}
  \centerfloat	
        \includegraphics[width=2.8\columnwidth]{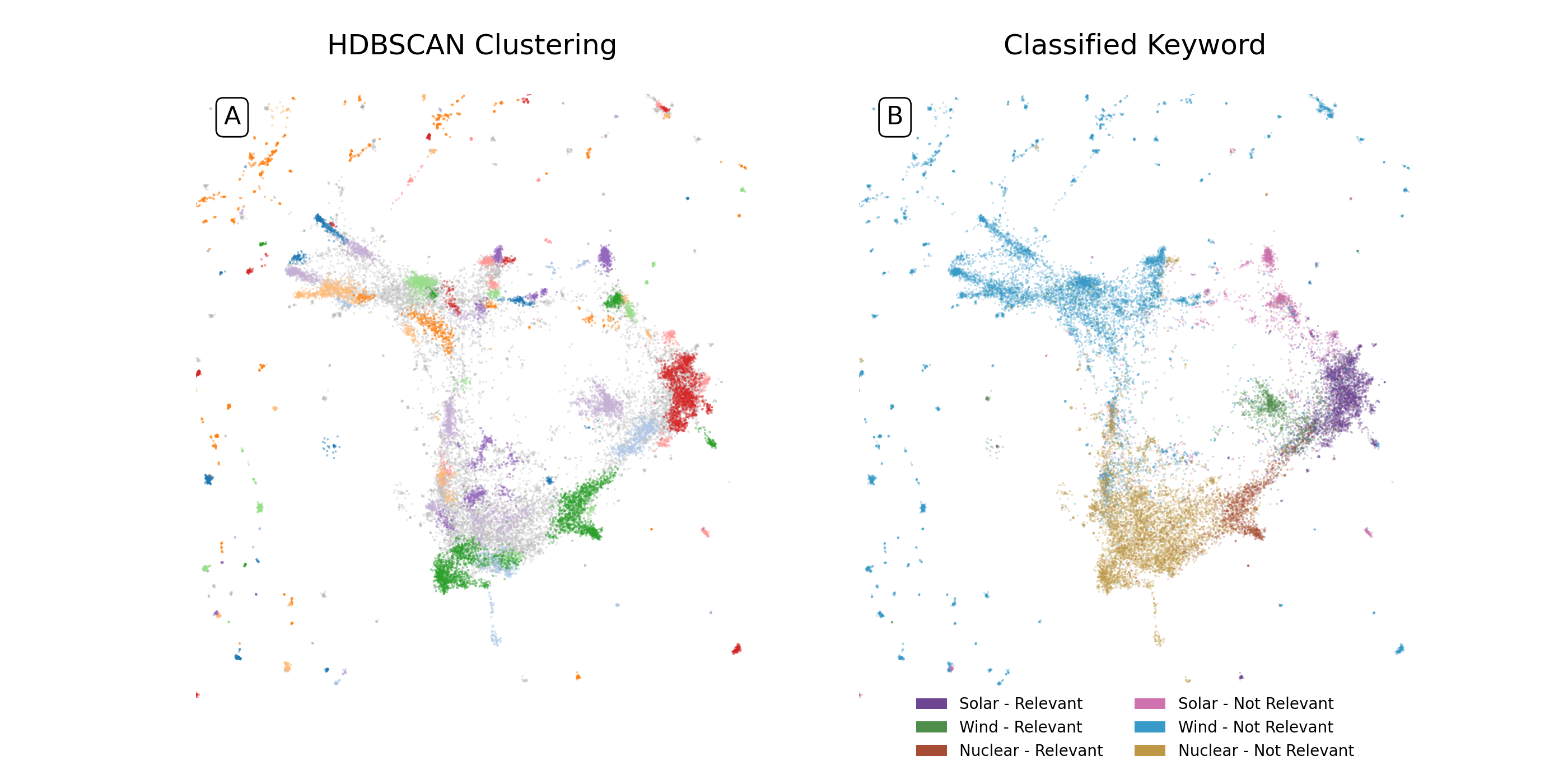} 
  \caption{
  \textbf{Embedded tweet distribution plot for the combined datasets.}
  Using a pre-trained model for semantically meaningful sentence embeddings based on MPNet, we plot the distribution of tweets within this semantic space.
  In both plots, points are tweets projected into 2D using UMAP for dimensionality reduction~\cite{mcinnes2018umap}.
  In panel A, we perform density based, hierarchical clustering using HDBSCAN and color by cluster.
  In panel B, we color by both the keyword used to query and the classification as relevant or non-relevant to the topic of clean energy. 
  Relevant tweets containing the keywords \anchor{wind}, \anchor{solar}, and, to a lesser extent, \anchor{nuclear} are relatively close together on the right in the embeddings, while non-relevant tweets are more dispersed.
  } 
    \label{fig:combined_embeddings}
\end{figure*}

\section{Results}
\label{sec:corpusCreation.results}

\subsection{Interpretations of sentence embeddings}
\label{sec:corpusCreation.results.embeddings} 

We first examine our corpus within a semantically meaningful sentence embedding, shown in Fig.~\ref{fig:combined_embeddings}. 
For each tweet,
we compute embeddings using \texttt{all-mpnet-base-v2},
a high performing, general-purpose sentence embedding model based on MPNet. 
The model is pre-trained to minimize cosign distance between a corpus of 1 billion paired texts 
and accessed using the sentence transformers python package~\cite{reimers-2019-sentence-bert}.

We include embeddings of all three corpora, anchored by the keywords \anchor{solar}, \anchor{wind}, and \anchor{nuclear}, 
and project onto two dimensions for visualization using Uniform Manifold Approximation and Projection (UMAP) for dimensionality reduction~\cite{mcinnes2018umap}. 
In the 2D projection, semantic distances between words are distorted.
Local relationships are preserved, but global position and structure is not. 

In Fig.~\ref{fig:combined_embeddings}A, we perform unsupervised clustering using HDBSCAN, a density  and color by cluster~\cite{mcinnes2017hdbscan}. 
Although we cannot share the interactive version of these plots, which allow the individual tweet texts to be read, we can summarize as follows. 
On the right side,
a large red cluster contains tweets that are primarily about solar energy. 
To the left in light blue, we identify a dense cluster of wind and solar tweets. 
Nearby in light purple, we find a cluster of wind energy related tweets.
The close green cluster contains nuclear energy tweets, with those being closer to the solar and wind tweets more likely to mention renewable energy source, while those further away only discuss nuclear in isolation.

We found the performance of the semantic embedding impressive, but clustering within this embedding was unsuitable for corpus curation. 
For example,
tweets arguing the relative merits of multiple technologies fell into a lower density location in the embedding space, and were classified as outliers by HDBSCAN, though they would clearly be classified as relevant by human raters. 

In Fig.~\ref{fig:combined_embeddings}B, we show the results of our three supervised text classifiers, based on MPNet trained for sentence embeddings and fine-tuned on a dataset of 1000 labeled tweets for each keyword.
The local positioning of tweets within the embedding reflects similarity in the sentence embedding space.
Tweets classified as relevant to clean energy technologies are clustered on the right-hand side, and overlap where they are mentioned together.
For paraphrased example tweets within each classification, refer to Tab.~\ref{tab:example_tweets}.

On the bottom third of the embedding, relevant \anchor{nuclear} tweets smoothly transition into non-relevant tweets, reflective of the occasionally blurry line between nuclear energy and weapons programs.

\anchor{Solar} tweets, by contrast, are easily separable. 
Phrases like `solar system', `solar eclipse', and `solar opposites' (a television sitcom) are common example usages.
These are entirely unrelated to solar energy
and the sentence embedding model places them in distinct regions of the semantic space.

Relevant \anchor{wind} tweets are also clearly separable from non-relevant tweets,
which often contain phrases related to the weather,
such as `wind storm' or `wind speed',
or more rhetorical expressions like `wind up' or `second wind'.
A number of weather bots regularly report wind speed measurements with a template format changing only speed and location. 
These tweets become close neighbors in the semantic embedding and, when projected onto two dimensions by UMAP, are split off from the larger connected component and pushed to the outer edge. 


\subsection{Ambient time series plots}
\label{sec:corpusCreation.results.ambient} 

For each case study we compare the text in the relevant corpus to the non-relevant corpus with three figure types. 
The first are ambient sentiment time series plots, shown in Figs. \ref{fig:solar_sentiment}, \ref{fig:wind_sentiment}, and \ref{fig:nuclear_sentiment}.
By sentiment we broadly mean the semantic differential of good-bad (or positive-negative).
In these plots we show dynamic changes in language use for tweets containing the selected anchor keyword over time.
On the top panel, we show the number of n-gram tokens with LabMT sentiment scores within each time bin~\cite{dodds2011temporal}.
In the center panel, we plot the ambient sentiment, $\Phi$, 
using a dictionary of LabMT sentiment values $\phi_{\tau}$. For each word $\tau$.
Wee compute the ambient sentiment as the weighted average, 
\begin{equation}
\Phi_{\textnormal{avg}}
= 
\sum_{\tau} 
\phi_{\tau}
p_{\tau},
\end{equation}
where $p_{\tau}$ is the probability or normalized frequency of occurrence. 
Error bars represent the standard deviation of the mean, with $N$ set conservatively as the number of tweets, rather than number of tokens.

In the lower panel, we plot the standard deviation of ambient sentiment,
which could help indicate when the distribution of sentiment is becoming narrower, broader, or even bimodal,
indicating polarization.
We plot three measurements for three corpora, 
tweets classified as relevant (R), non-relevant (NR), and the combined dataset (R + NR),
with the latter reflecting the measurements we would have obtained without training a classifier.

\subsection{Lexical calculus: Word shift plots}
\label{sec:corpusCreation.results.wordshifts} 

To examine how the average sentiment differs between the relevant and non-relevant corpora,
we present three sentiment shift plots in 
Fig.~\ref{fig:combined_sentiment_shifts}~\cite{gallagher2021generalized}.
Word shifts allow us to visualize how words individually contribute
to differences in average sentiment between two texts, a reference and a comparison text.
Words that contribute to the comparison text having a higher sentiment than the reference, are shown having a positive contribution, $\delta \Phi_{\tau}$. 
Bars corresponding to words with a higher rated sentiment score than the average of the reference text are colored yellow, or blue if lower. 
Finally, we rank words by the absolute value of their contribution to the difference in average sentiment, 
$\delta \Phi_{\textnormal{avg}}$, 
giving a list of the top contributing words.

\subsection{Allotaxonometry}
\label{sec:corpusCreation.results.allotax} 

We further compare language usage using an allotaxonograph in Fig.~\ref{fig:rankdiv_solar}, 
an interpretable instrument that provides a rank-rank histogram of word usage
and a ranked list of rank-turbulence divergence (RTD) contributions from individual words. 
Being able to compare the 1-gram or 2-gram distributions of two corpora with RTD allows us to extract characteristic words at all scales~\cite{dodds2020allotaxonometry}.
To compute RTD, 
we take each distinct word, $\tau$, and compute the ranks with each corpus, $r_{\tau, 1}$ and $r_{\tau, 2}$. 
RTD is the sum the difference between inverse ranks, scaled with a parameter, $\alpha$, and normalized to lie between 0 and 1, having the form:
\begin{equation}
D_\alpha (R_1 \| R_2)  
\propto 
\sum   
\left\lvert
  \frac{1}{\left[r_{\tau,1}\right]^{\alpha}}
  -
  \frac{1}{\left[r_{\tau,2}\right]^{\alpha}}
\right\rvert^{1/(\alpha+1)}.
\end{equation}
We set $\alpha = 1/4$ for social media corpus comparisons~\cite{dodds2020allotaxonometry}.

We intend that the following cases studies may serve as an example set of procedures and provide diagnostic tools for computational social scientists to adopt this approach to social media corpus curation.

\subsection{Solar Energy Case Study}
\label{sec:corpusCreation.results.solar} 

\begin{figure}[tp!]
  \centering	
    \includegraphics[width=0.98\columnwidth]{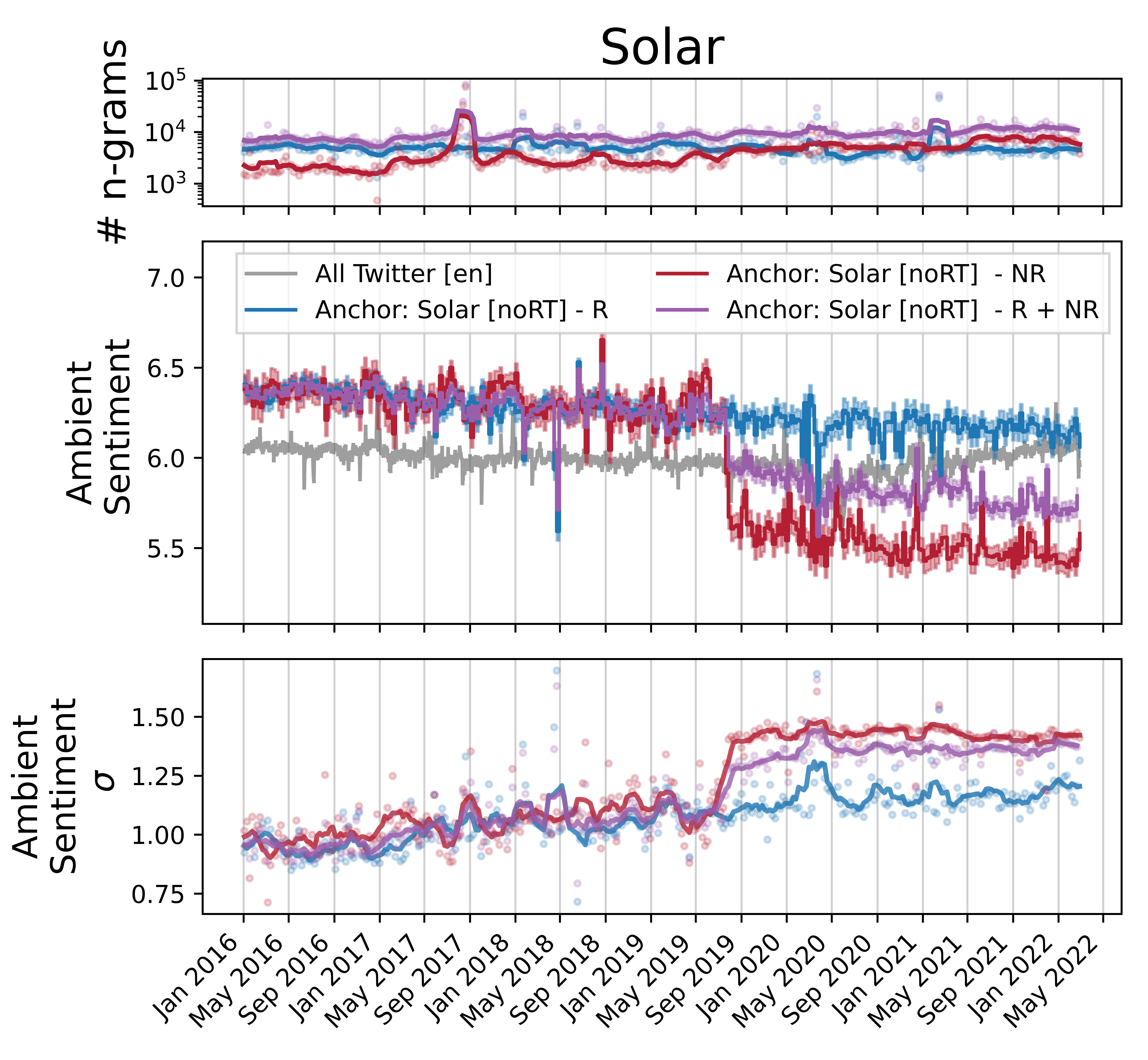}  
  \caption{
    \textbf{Ambient sentiment time series comparison for relevant  (R), non-relevant (NR), and combined tweet corpora, containing the keyword \anchor{solar}.}
    In the top panel, we show the number of tokens with LabMT \cite{dodds2015human} sentiment scores in each corpus on each day.
    `Relevant' tweets, in blue, have more scored tokens early on,
    but the number tokens in `non-relevant' tweets increase in relative proportion over time.
    The center panel shows the average sentiment for each corpus, including a measurement of English language tweets as a whole in gray for comparison. 
    Before 2019, the measured sentiment for both corpora are comparable, but subsequently the mean sentiment of `non-relevant' tweets drops. 
    In the bottom panel we plot the standard deviation of the sentiment measurement, which captures a broader distribution of sentiment scores for  `non-relevant' tweets.
    Without classification filtering, the ambient sentiment measurement would be entirely misleading, appearing as though the sentiment contained in tweets containing the word \anchor{solar} dropped dramatically in 2019, when in fact sentiment has only modestly declined. 
  }
  \label{fig:solar_sentiment}
\end{figure}

Solar tweets were nearly evenly split with 47\% of the corpus being relevant and 53\% being non-relevant by volume of words.
The solar tweet corpus also achieved the highest classification performance with an F1 score of 0.95, 
as shown in Tab.~\ref{tab:F1-scores}.

Of the three case studies, 
we find the R \anchor{solar} tweets corpus evolves
most relative to the corresponding NR corpus.
Looking at the sentiment time series 
in Fig.~\ref{fig:solar_sentiment}, 
we see little difference between the ambient sentiment of the R and NR corpora prior to 2019.

In May of 2019, NR ambient sentiment, shown in red, sharply falls while the R corpus appears to remain on trend.
For the standard deviation of ambient sentiment, 
which measures the width of the distribution of sentiment scores for each LabMT word in the ambient corpus,
we also observe a dramatic increase in 2019.

We find that this shift in language use in the NR corpus occurs without a change in query terms, 
and demonstrates how simple keyword queries can fail. 
We contend that the process of selecting relevant social media documents to include in a corpus
is just as important as the NLP measurement tools
used to quantify sentiment.
The difference in resulting sentiment measurements,
between what would have been measured without a classifier
(the R + NR corpus in purple) 
and the improved measurement after filtering with a classifier 
(the R corpus in blue) 
is stark. 
Looking at only the combined R + NR measurement,
researchers could incorrectly conclude that language surrounding \anchor{solar}
has decreased in sentiment dramatically since 2019. 

Focusing on only the R \anchor{solar} sentiment time series,
we see clearly that there was in fact
no dramatic drop in sentiment around \anchor{solar}, 
and the relevant language around solar 
remains more positive relative to English language tweets in general. 
The decrease in observed NR sentiment is related to an influx of weather bots,  
which provide updates as often as hourly on local weather conditions
and contain \anchor{solar} used in the context of measuring current solar radiation. 
In Fig.~\ref{fig:combined_sentiment_shifts} we see terms like `radiation', `pressure', and `humidity' are contributing to a lower average sentiment for the NR corpus.

\begin{figure*}
  \centering	
    \includegraphics[width=0.90\textwidth]{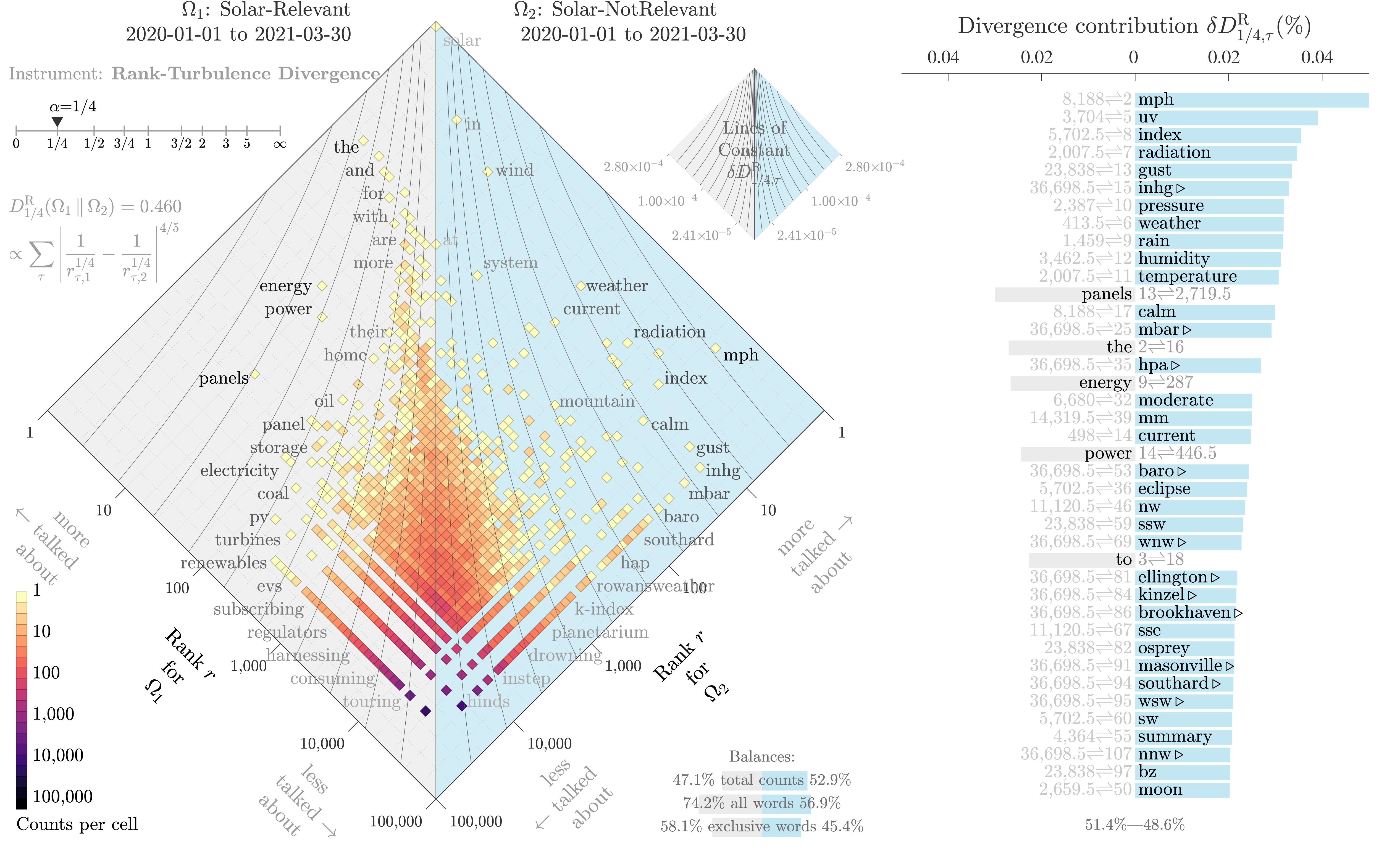}  
  \caption{
  \textbf{Allotaxonograph comparing the rank divergence of words
    classified as relevant to solar energy discourse
    to those containing the keyword \anchor{solar}
    but classified as non-relevant.} 
    On the main 2D rank-rank histogram panel,
    words appearing on the right have a higher rank in the `relevant' subset than in `non-relevant', while phrases on the left appeared more frequently in the `non-relevant' tweets.
    The panel on the right shows the words which contribute most to the rank divergence between each corpus.
        We observe that many words associated with weather bots,
    such as `mph,' `uv,' and `pressure,' 
    are more frequently used in non-relevant posts,
    while words like `panels,' `energy,' and `power,' used more in tweets relevant to solar energy.
    Notably, commonly used function words,
    such as `the,' `and,' and `are,' 
    are off-center in the rank-rank histogram, a further indication that many of the `non-relevant' tweets are from automated accounts publishing weather data rather than using conversational English.
    The balance of the words in these two subsets is noted in the bottom right corner of the histogram, showing the percentage of total counts, all words, and exclusive words. 
    For this example the two subsets are nearly balanced, indicating that the filtered corpus contains less than 50\% of word tokens from the raw query.
    See Dodds~\etal~\cite{dodds2020allotaxonometry} for a full description of the allotaxonometric instrument.}
    \label{fig:rankdiv_solar}
\end{figure*}

Examining the rank-turbulence divergence shift for \anchor{solar} from January 2020 to March 2021 in 
Fig.~\ref{fig:rankdiv_solar}, we can see terms like `energy', `power', and `panels' are much more common in the R corpus, all being among the top 15 most frequently used terms. 
On the other side of the ledger,
we find weather related terms like `mph', `uv', `radiation', and `gust' to be top words in the NR corpus. 
We also observe that function words---e.g., `the', `to', and `for'---are more common in the R corpus, skewing the rank-rank histogram to the left. 
The lack of function words is another result of weather bots dominating in the latter period of our study.

\subsection{Wind Energy Case Study}

\begin{figure}[tp!]
  \centering	
    \includegraphics[width=0.98\columnwidth]{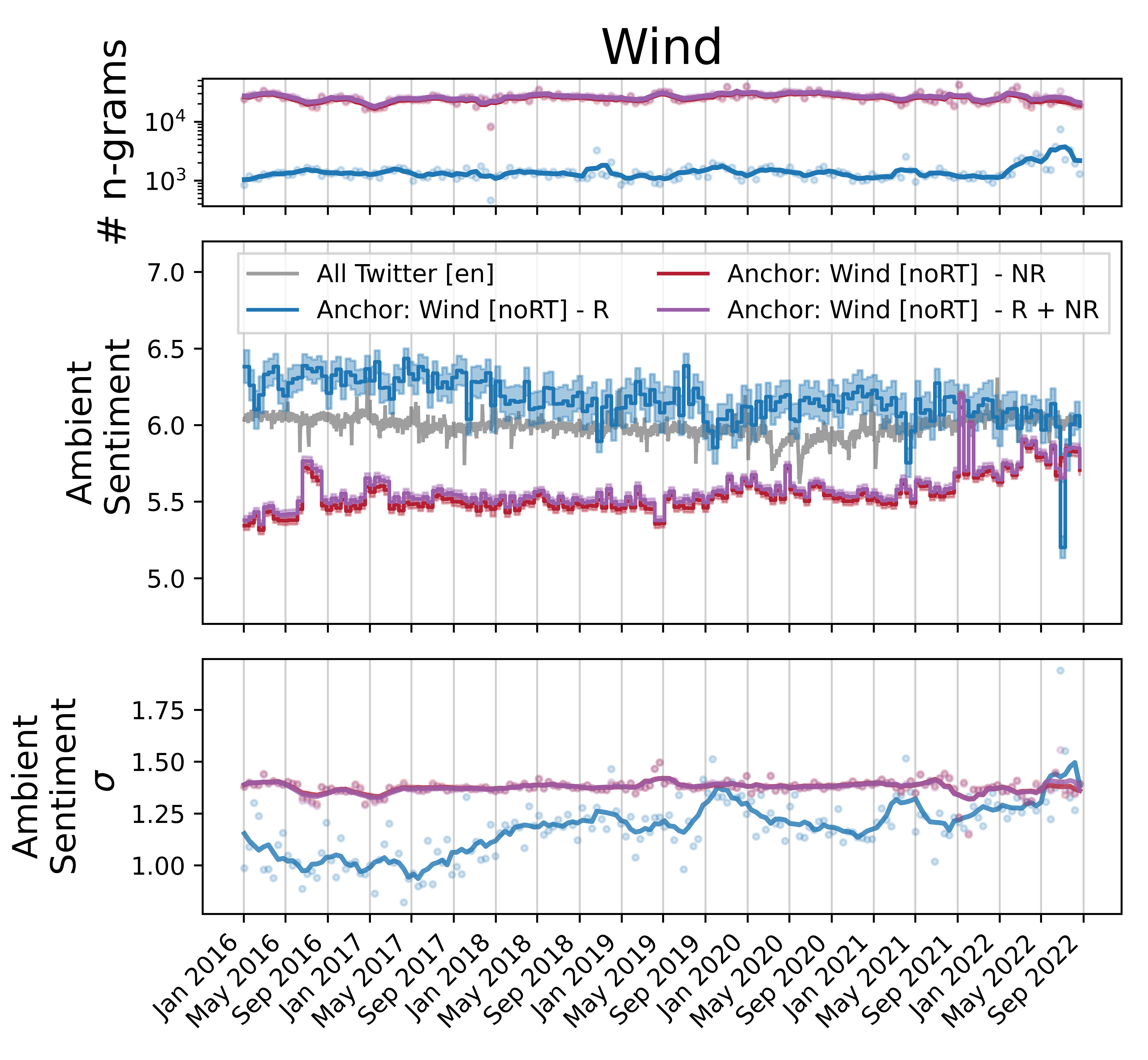}  
  \caption{
    \textbf{Ambient sentiment time series comparison for relevant  (R),
    non-relevant (NR), and combined tweet corpora,
    all containing the keyword \anchor{wind}.}
    In the top panel,
    we show the number of tokens with LabMT
    sentiment scores for each corpus during each two week period~\cite{dodds2015human}.
    R tweets, in blue, have more than an order of magnitude fewer tokens per time window over the entire study period.
    The center panel shows the average sentiment for each corpus, including measurement of English language tweets as a whole in gray for comparison. 
    R \anchor{wind} tweets are more positive than Twitter on average early on, but this difference is reduced over time. 
    Because most \anchor{wind} tweets are non-relevant, sentiment of the combined corpus closely follows the NR sentiment.
    In the bottom panel we plot the standard deviation of the sentiment measurement,
    which captures a broader distribution of sentiment scores for  `non-relevant' tweets,
    as was the case for all case-studies we examined.
    Without classification filtering, the ambient sentiment measurement would have been dominated by NR tweets. 
  }
  \label{fig:wind_sentiment}
\end{figure}

The unclassified \anchor{wind} tweets corpus had the lowest proportion of relevant tweets.
Only 5\% of the human labeled subset was related to clean energy.
The $n$-gram \anchor{wind} is used in many different contexts besides energy generation,
from casual discussion of today's weather to figurative uses like references to athletes getting their `second wind' and the anticipatory rotational phrase `wind up' where `wind' rhymes with `kind'.
In the top panel of Fig.~\ref{fig:wind_sentiment},
we see that the number of n-grams in relevant tweets
with corresponding sentiment scores is consistently around $10^3$, 
while the NR corpus 
contains more than an order of magnitude more text.

We found the ambient sentiment of the R \anchor{wind} corpus has been slightly more positive than average language use on Twitter. 
The NR corpus had distinctly lower sentiment, but is more dynamic, rising from a low of 5.5 in 2016, to 5.9 in 2020.
Because the proportion of tweets relevant to energy is so low, 
the combined sentiment time series measurement is dominated by the NR corpus. 
The standard deviation of sentiment, $\sigma$, for the R corpus also increases from around 1.0 in 2016, before leveling off around 1.2, slightly under the NR corpus.

The choice of \anchor{wind} could seem to be a
poor choice of keyword, 
given that the vast majority of matching tweets are non-relevant.  
Under a paradigm  of expert-crafted lists of keywords, we would indeed agree such a generously matching term 
would not be suitable. 
However, by choosing a potentially ambiguous term, 
we are able to capture a wider range of users. 
Those who do not wish to project their thoughts into a global conversation by attaching a hashtag, 
but are content with discussing among their local network, are still included with this methodology. 
Also included are users writing informally or using context of a threaded conversation, who might not use a high precision keyword phrase, like `wind power', `wind generation', or `wind energy'.
These cases make up a significant proportion of conversation around any given topic; researchers studying more obscure topics could benefit from the increased sample size, and temporal resolution of a higher recall set of keywords.

\subsection{ Nuclear Energy Case Study}

\begin{figure}[tp!]
  \centering	
    \includegraphics[width=0.98\columnwidth]{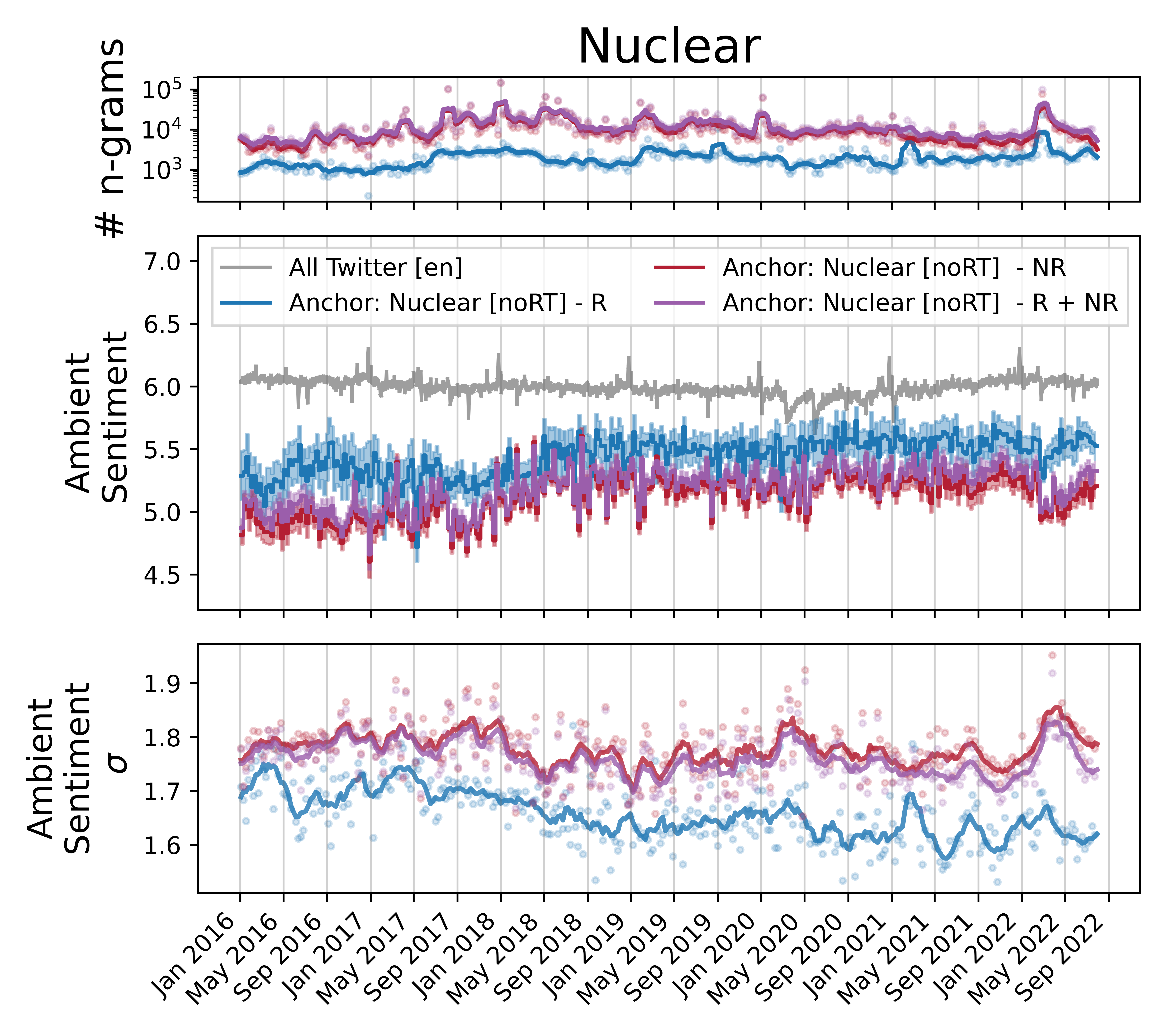}  
  \caption{
    \textbf{Ambient sentiment time series comparison for relevant (R), non-relevant (NR), and combined tweet corpora,
    all containing the keyword \anchor{nuclear}.}
    In the top panel,
    we show the number of tokens with LabMT \cite{dodds2015human}
    sentiment scores for each corpus in each two week period.
    The number of relevant n-grams,
    in blue,
    is consistently lower than non-relevant n-grams. 
    The center panel shows the average sentiment for each corpus,
    including measurement of English language tweets as a whole in gray. 
    We found that
    R tweets had higher sentiment than NR tweets
    containing \anchor{nuclear},
    but had much lower sentiment than Twitter as a whole.
    Sentiment appears relatively stable for both corpora
    with periods of higher sentiment around 2017 and 2020-2022 for the R corpus.
    In the bottom panel, we plot the standard deviation of the sentiment measurement, which shows a broader distribution of sentiment scores for NR tweets,
    as well as sentiment for both corpora trending down slightly. 
  }
  \label{fig:nuclear_sentiment}
\end{figure}

The \anchor{nuclear} case study had the lowest classification performance after fine-tuning, 
achieving an F1 score of 0.86. 
The proportion of relevant tweets, 16\%, was higher than for the \anchor{wind} corpus. 
We believe the performance was impacted negatively by the close proximity and overlap
of nuclear energy and nuclear weapons topics in the semantic embedding space. 

The ambient sentiment time series, 
in Fig.~\ref{fig:nuclear_sentiment},
for the R \anchor{nuclear} corpus was much lower than average sentiment on Twitter for the entire study period,
but higher than the NR corpus.
It appears that ambient sentiment around R nuclear energy tweets has been increasing, with a higher stable level since fall 2020. 
We found that the standard deviation of sentiment is also decreasing slightly,
though it starts from a much higher level of around 1.7, when compared with wind and solar.

\begin{figure*}
  \centering	
    \includegraphics[width=0.98\textwidth]{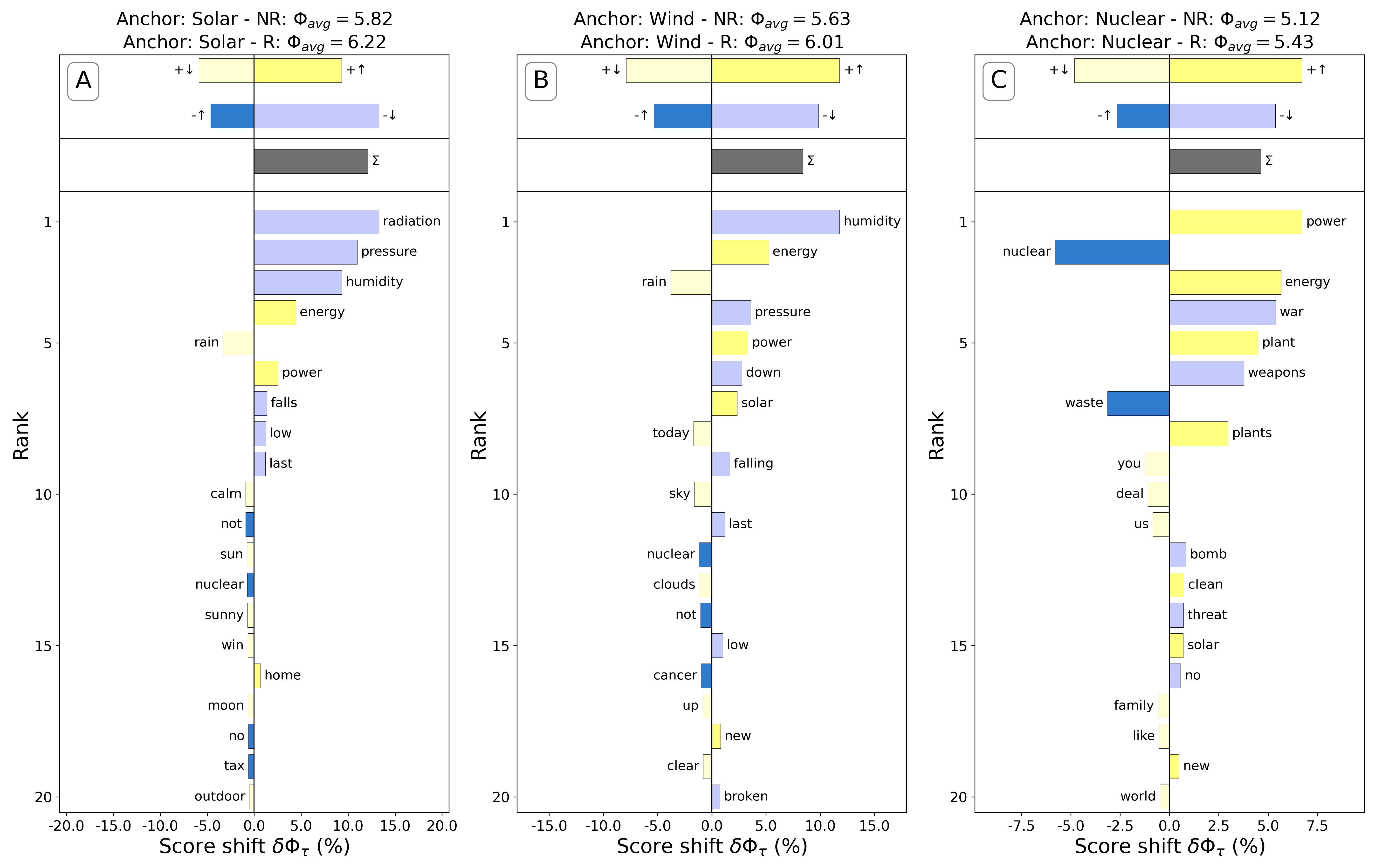}  
  \caption{
    \textbf{Sentiment shift plots comparing the classified relevant (R) and non-relevant (NR) tweet corpora for tweets containing the keywords \anchor{solar}, \anchor{wind}, and \anchor{nuclear}.}
    We show the top 20 words contributing to the difference in LabMT sentiment between the corpora.
    \textbf{A.}
    Relevant tweets that are related to clean energy
    are more positive on average for all keywords
    when compared to non-relevant tweets. 
    Sad words that are less common in relevant \anchor{solar} tweets are `radiation', `pressure', and `humidity', which largely refer to the weather.
    Happy words like `energy' and `power' are more common in relevant tweets compared to tweets non-relevant to solar energy. 
    \textbf{B.}
    For \anchor{wind}, relatively sad terms like `humidity' and `pressure' are less common in relevant tweets
    (these appear in clearly non-related tweets about the weather),
    while happy terms like `energy', `power', and `solar' are more common in tweets relevant to wind as a renewable energy source. 
    \textbf{C.}
    For \anchor{nuclear}, relevant tweets are on average more positive due to sad words like `war', `weapons', and `bomb' being less common in relevant tweets,
    while happy words like `power' and `energy' are more common.
    The two prominent sad words `nuclear' and `waste'
    go against the positive difference in moving
    from non-relevant to relevant tweets as they 
    both occur more frequently in relevant tweets.
  }    
  \label{fig:combined_sentiment_shifts}
\end{figure*}

In Fig.~\ref{fig:combined_sentiment_shifts}, 
we can see that the \anchor{nuclear}
R corpus's higher sentiment
relative to that the NR corpus
is driven by more positive words like `power' and `energy',
but also fewer negative words,
like `war' and `weapons'.
Going against the grain is the word `nuclear' itself
as well as term `waste' 
which are both negatively scored words that are
used much more frequently in the R corpus 
relative to the NR corpus.

\section{Concluding remarks}
\label{sec:corpusCreation.concludingremarks}

Disambiguating relevant tweets has been a challenge for researchers,
especially when a natural keyword choice has a commonly used homograph~\cite{ginart2016drugs}.
We have demonstrated that text classifiers can be trained on top of pre-trained contextual sentence embeddings,
which can accurately encode researcher discretion
and infer the relevance of millions of messages on a laptop. 

Rather than defining the boundaries of a corpus by a set of expert chosen keywords or expert crafted query rules,
researchers can look at a sample of data, 
label messages as relevant as they see see fit,
and communicate their reasoning directly. 
Reviewers and skeptical readers would be empowered to make their own judgments
of what qualifies as a relevant tweet, 
by labeling themselves and comparing the resulting text measurements.

Classification for social media datasets is not a panacea; Twitter's user base remains a non-representative sample of populations, skewing younger, more male, and more educated~\cite{mislove2011understanding}. 
A small proportion of prolific users generate an outsized proportion of text, while most users rarely tweet~\cite{wojcik2019sizing}.
Despite these problems, the platform remains a critical source of data on public conversations at the time of writing 
with a low barrier to entry compared to traditional media.

Future work could explore better sampling methods for humans labeling tweets
to reduce the amount of labeled data needed to train the text classifier. 
Sampling messages by shuffling
risks oversampling
from dense regions of the semantic embedding space. 
The coder sees repetitive messages that provide little marginal 
information to the model.
This would have negative impacts on the generalizability of the classifier, and we would be skeptical of real-time measurements as conversation could drift into under-explored regions of the semantic embedding space.
Other work could explore the trade-offs between optimizing for high recall and high precision when curating  social media datasets, and the impacts on resulting measurements. 

For online applications of relevance classifiers, such work would be useful in identifying when more training data is needed. 
By measuring changes in language use, 
both by measuring rank-turbulence or probability-turbulence divergence ~\cite{dodds2020allotaxonometry,dodds2020g}
between the training corpus and incoming data,
and by measuring changes in the distribution of messages within a semantic embedding, thresholds for train data updates could be determined.

Finally, researchers could explore viewing social media datasets as having uncertain boundaries, 
and running measurements over data set ensembles
to better capture the uncertainly in researcher discretion inherent in corpus curation.

Overall, we hope our work here highlights
a viable alternative corpus curation method
for computational social scientists studying social media datasets.

\acknowledgments
This material is based upon work supported by the National
Science Foundation under Award No. 2242829. 
The authors are also grateful for 
support furnished by MassMutual and Google,
and the computational facilities provided
by the Vermont Advanced Computing Center.

\clearpage

\end{document}